%% file: root.tex
\documentclass[letterpaper, 10 pt, conference]{ieeeconf}  

\IEEEoverridecommandlockouts                              

\usepackage{multirow}
\usepackage{lipsum}
\usepackage{amsmath}
\usepackage{amssymb}
\usepackage{siunitx}
\usepackage{bm}
\usepackage{graphicx}
\usepackage{subcaption}
\usepackage{booktabs}
\usepackage{comment}
\usepackage[acronym]{glossaries}
\makeglossaries
\loadglsentries{gls}
\usepackage[ruled,vlined]{algorithm2e}
\usepackage{graphicx}
\graphicspath{{./Figures/}}
\usepackage{xurl}
\usepackage[hidelinks]{hyperref}
\input{math_commands}

\usepackage[english]{babel}
\title{
\LARGE \bf Physics-Regularized Machine Learning for Proprioceptive \\Vehicle Localization Using Onboard Sensors}

\author{
	Abinav Kalyanasundaram$^{1}$, Karthikeyan Chandra Sekaran$^{1}$, Wolfgang Utschick$^{2}$ and  Michael Botsch$^{1}$
    \thanks{$^{1}$AImotion Bavaria, Technische Hochschule Ingolstadt, Germany, {\tt\small firstname.lastname@thi.de}}%
    \thanks{$^{2}$Technische Universität München, Germany, {\tt\small utschick@tum.de}}%
}

\begin{document}
\bstctlcite{IEEEexample:BSTcontrol}
\maketitle
\thispagestyle{empty}
\pagestyle{empty}

\begin{abstract}
Accurate and robust localization is essential for autonomous mobility systems in real-world environments.
While fusing \gls{imu} data with satellite-based correction signals provides precise vehicle pose estimates, performance degrades substantially during outages.
Recent studies indicate that \gls{ml} can improve \gls{imu}-based proprioceptive localization, highlighting untapped potential for onboard sensors readily available in production vehicles. 
This paper introduces~\gls{piml2}, a hybrid framework that combines the complementary strengths of Kalman filtering and data-driven learning to estimate vehicle pose directly from onboard sensors.
A key aspect of \gls{piml2} is its physics-regularized learning, enabled by end-to-end training of an \gls{ml} model through a differentiable Kalman filter. 
This improves consistency with vehicle motion models, thereby enhancing both localization accuracy and generalization across driving conditions. 
We evaluate the performance limits of \gls{ml}-enhanced onboard odometry on a publicly available dataset and show that \gls{piml2} achieves superior localization accuracy and demonstrates real-time capability. 
This work also introduces a novel dataset to support vehicle localization research under low-friction conditions. 
The proposed framework provides a robust and cost-effective solution for vehicle localization under degraded sensing conditions by integrating learning with physics-based priors.


\end{abstract}

\section{Introduction}
Precise localization is crucial for the safe and reliable operation of autonomous vehicles. The knowledge about the ego vehicle's pose acts as a foundation for critical downstream tasks such as path planning, motion control, and decision-making~\cite{intro_localization}. Accurate localization is typically achieved by fusing an \gls{imu} with \gls{gnss} measurements~\cite{gnss}. However, during \gls{gnss} outages, dead reckoning based on commercial-grade \gls{imu}s rapidly degrades due to drift accumulation~\cite{lstmIMU}. While fusing \gls{imu}s with cameras and LiDAR can reduce drift, such approaches increase system cost, complexity, and are sensitive to environmental conditions~\cite{survey_localization}. Alternatively, commonly available onboard sensors such as steering wheel angle, wheel speeds, and yaw rate provide valuable odometry signals for vehicle localization~\cite{abinav,sven}. Being proprioceptive, they can measure vehicle motion independently without reliance on external signals or the environment. Therefore, exploring the achievable performance of vehicle localization using only onboard sensors is critical for robust and cost-effective autonomy under degraded \gls{gnss} conditions.


\gls{sota} methods for vehicle localization during \gls{gnss} outages can be broadly categorized into three groups: HD map–based methods, multi-sensor fusion techniques, and dead-reckoning approaches~\cite{localisation_survey}. HD map–based methods achieve high localization accuracy but require pre-built and continuously maintained maps~\cite{survey_localization}. Multi-sensor fusion techniques, such as VINS, LiDAR-IMU odometry, and SLAM, often require expensive sensors and can be sensitive to environmental conditions~\cite{survey_localization,imu_survey}. In contrast, dead reckoning based on \gls{imu}s or onboard sensors is proprioceptive and inexpensive, yet prone to drift~\cite{imu_survey}. Classical Bayesian filters for \gls{imu}s generalize well but are limited by noise accumulation, leading to significant drift during \gls{gnss} outages~\cite{lstmIMU,noise-imu-degrade}. Data-driven learning methods can tackle sensor noise but suffer from a lack of physical consistency, generalization across driving conditions, and challenges in real-time deployment~\cite{noise_imu}. 
These complementary limitations motivate hybrid \gls{ml} approaches that combine the generalization and physical consistency of model-based filtering with the noise-handling capacity of data-driven methods.
While \gls{ml}-based \gls{imu} drift correction and inertial odometry have been extensively investigated~\cite{rnn-ekf,dl-avl}, the potential of \gls{ml}-enhanced odometry using standard onboard vehicle sensors remains relatively underexplored~\cite{onboard-sensor}.
\begin{figure}[tb]
    \centering
        \includegraphics[scale=1.15, trim=10pt 10pt 10pt 10pt, clip]{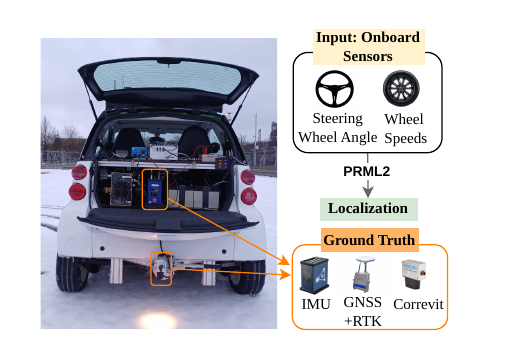}
    \caption{Overview of the test vehicle configuration and the objective of \gls{piml2} for proprioceptive localization.}
    \label{fig:piml_architecture}
    \vspace{-0.5cm}
\end{figure} 

To overcome these limitations, this work introduces \textbf{P}hysics-\textbf{R}egularized \textbf{M}achine \textbf{L}earning for
\textbf{L}ocalization (PRML2), a novel hybrid architecture for vehicle localization that relies solely on onboard sensors as illustrated in Fig.~\ref{fig:piml_architecture}. \gls{piml2} integrates a transformer-based~\cite{attention}~\gls{ml} model with a physics guard layer and an adaptive~\gls{ekf}, as shown in Fig.~\ref{fig:informerarchitecture}. 
A key feature of \gls{piml2} is the end-to-end training of the \gls{ml} model through the differentiable \gls{ekf}, which acts as a physics-informed regularizer to improve vehicle dynamic consistency and generalization. By relying exclusively on onboard sensors, \gls{piml2} is cost-effective, always available, and reduces drift by integrating learning with physics-based filtering. The main contributions of this work are as follows:

\begin{itemize}
\item Introduction of \gls{piml2}, an end-to-end trainable hybrid architecture integrating a transformer-based \gls{ml} model with a differentiable \gls{ekf} for vehicle localization using onboard sensor measurements.
\item Theoretical and empirical validation that Kalman filters act as a physics regularizer in end-to-end training, penalizing physical and temporal inconsistencies.
\item Demonstration of \gls{piml2}’s ability to generalize across challenging low-friction driving conditions.
\item Publication of a novel dataset to support vehicle localization research under low-friction conditions.\footnote{Dataset and code to replicate results are available at \url{https://github.com/MB-Team-THI/PRML2-for-Vehicle-Localization}}

\end{itemize}
\section{Related works}

Proprioceptive sensors measure a vehicle’s motion directly, without relying on external signals or environmental features. While \gls{imu}s are the most common proprioceptive sensors used for localization~\cite{localisation_survey}, production vehicles also contain standard onboard sensors that provide high frequency motion-related information~\cite{abinav,onboard-sensor}. Proprioceptive localization methods can be broadly categorized into model-based Bayesian filtering approaches and data-driven \gls{ml} methods.


Recursive Bayesian filters, including the \gls{kf}, \gls{ekf}, error-state \gls{ekf}, and particle filters, perform dead reckoning by propagating motion models and fusing IMU measurements~\cite{survey_localization}. 
Although these methods generalize well, they are vulnerable to complex, multifaceted noise that degrades localization accuracy over time~\cite{lstmIMU,noise-imu-degrade}.
Adaptive filtering strategies attempt to mitigate this by adjusting process and measurement noise covariances~\cite{rnn-ekf,adaptive_kalman}, yet \gls{imu}-based dead reckoning remains prone to quadratic growth of position error~\cite{lstmIMU}.
Incorporating velocity information, such as wheel-mounted \gls{imu} measurements~\cite{wheel-imu}  or onboard sensor-based velocity estimates~\cite{onboardsensors-velocity} using vehicle dynamic models, can reduce drift rate. However, complete onboard sensor-based localization is fundamentally constrained by the assumptions of the vehicle motion models, which hold only under certain driving conditions~\cite{icra_wheelodometry}. 



Data-driven proprioceptive localization approaches employ \gls{ml} techniques to calibrate \gls{imu} measurements~\cite{imu_calibrate}, compensate integration drift~\cite{deep_position}, and adaptively model sensor noise~\cite{noise_imu,adaptive_noise_israel}.  In~\cite{calibnet2022}, deep learning was used to enhance the performance of low-cost \glspl{imu} and mimic the output characteristics of high-grade sensors. Although \gls{ml}-based methods have been widely explored for handling stochastic noise in \gls{imu}-based localization, their application to onboard odometry has been less studied. Recent works~\cite{abinav,dl-avl,TUdelft} demonstrate the potential of combining onboard sensors with \gls{ml} techniques, showing that \gls{ml} models can transform onboard sensor measurements into accurate vehicle dynamic state estimates without complex vehicle dynamic models. However, the main challenges of learning-based odometry include limited generalization and lack of compliance with physical constraints~\cite{imu_survey,noise_imu,nn-vdm}. 

Recently, differentiable filtering approaches~\cite{backpropkf,particlefilter,iros-diff} aim to bridge model-based estimation and data-driven learning by training neural networks with Bayesian filters. However, these methods have primarily been investigated in monocular visual odometry. We extend this paradigm to onboard proprioceptive localization by introducing a transformer-based measurement model with physics guard and provide insight into its physics regularization effect. 
To the best of our knowledge, real-time localization relying exclusively on standard onboard sensors, without IMUs or exteroceptive measurements, remains underexplored. 
We address this gap with \gls{piml2}, a framework orthogonal to multi-sensor fusion yet extensible to include additional sensing modalities.


\section{methodology}

\begin{figure*}[htbp]
\centering
\includegraphics[scale=0.91, trim=20pt 5pt 20pt 5pt, clip]{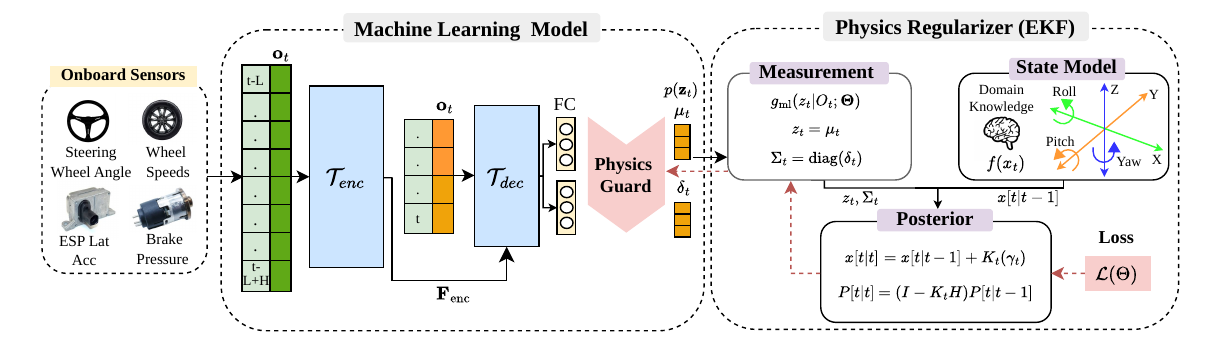}
\caption{The \gls{piml2} architecture contains two core components: A trasformer based \gls{ml} model and a differentiable \acrfull{ekf}. The \gls{ml} model processes the vehicle's onboard sensor measurements $\vo_t$ to estimate the vehicle dynamic state $\vz_t$ along with its uncertainty $\delta_z[t]$. A physics guard layer bounds these estimates within the vehicle's dynamic limits. Finally, these estimates act as pseudo measurements for the \gls{ekf} to compute the \textit{a posteriori} vehicle pose. In \gls{piml2}, the loss for training the \gls{ml} model is backpropagated through the differentiable \gls{ekf}, which acts as a physics  regularizer.}
\label{fig:informerarchitecture}
\vspace{-0.3cm}
\end{figure*}

This section presents the problem formulation and the proposed \gls{piml2} architecture. The overall framework is depicted in Fig.~\ref{fig:informerarchitecture}, and the subsequent subsections provide a detailed explanation about each component.
\subsection{Problem Formulation}
This work addresses the problem of vehicle localization using only onboard sensors in \gls{gnss}-denied conditions. 
The objective is to compute the dead-reckoning pose $\mT_k \in SE(3)$ at any time step $k$ in global coordinates, with $SE(3)$ denoting the space of 3D rigid-body transformations.
It can be recursively computed using the relative pose transformations $\mT_{t,t-1}$  from time \mbox{$(t-1)\!\to\!t$} as follows:
\begin{equation}
\label{Eqn:problem formulation}
\mT_k =  \Big(\prod_{t=1}^{k} \mT_{t,t-1}\Big) \mT_0 = \mT_{k,k-1} \cdots \mT_{1,0} \mT_0 ,
\end{equation}
where $\mT_0$ denotes the initial vehicle pose in global coordinates. 
This work proposes a physics-informed mapping function $g_{\boldsymbol{\Theta}}$ to compute the relative transformation from \gls{osd}: 
\begin{equation}
\label{function_transformation}
    \mT_{t,t-1}  = g_{\boldsymbol{\Theta}}(\mO_t ; L,m),
\end{equation}
where \mbox{$\mO_t = \{\mathbf{o}_{t-L+1}, \dots, \mathbf{o}_t\} \in \mathbb{R}^{L \times m}$} denotes the \gls{osd} for an observation window of $L$ time steps. 
Each measurement vector \ensuremath{\mathbf{o}_t = \{ o_t^j \}_{j=1}^{m} \in \mathbb{R}^m} 
contains the readings from $m$ onboard sensors at time step $t$. The data obtained from a precise vehicle localization system serves as a reference for ground truth (gt) vehicle state:
\begin{equation}
\label{ground_truth}
\vx_{t}^{\text{gt}} = 
\left[
\vd_{t}^{\text{gt}}, \boldsymbol{\theta}_t^{\text{gt}},\vz_t^{\text{gt}}\right]^{\top},
\end{equation}
where $\mathbf{d}_{t}^{\mathrm{gt}}$ and $\boldsymbol{\theta}_{t}^{\mathrm{gt}}$ denote the vehicle position and orientation, respectively, forming the vehicle pose. 
The vector $\mathbf{z}_{t}^{\mathrm{gt}}$ contains the vehicle dynamic quantities, including velocities $\mathbf{v}_{t}^{\text{gt}}$, accelerations $\mathbf{a}_{t}^{\text{gt}}$, and angular rates $\boldsymbol{\omega}_{t}^{\text{gt}}$.


\subsection{\gls{piml2} Architecture}
\label{sec:ml model}
The proposed~\gls{piml2} architecture, illustrated in Fig.~\ref{fig:informerarchitecture} represents the mapping function $g_{\boldsymbol{\Theta}}$ in Eq.~(\ref{function_transformation}) by decomposing it into two complementary components: a data-driven~\gls{ml} model and a filter-based~\gls{ekf}. The \gls{ml} model \mbox{$g_{\text{ml}}(\vz_t|\mO_t;\boldsymbol{\Theta})$} acts as a time-series regressor that maps onboard sensor measurements to vehicle dynamic states $\vz_t$, which the~\gls{ekf} uses to compute the pose. Since the proposed~\gls{ml} model performs sequence-to-sequence regression, a transformer-based architecture~\cite{attention} was chosen to capture long-range temporal dependencies in onboard sensor data, inspired by prior attention-based designs~\cite{abinav}.


\subsubsection{Machine Learning Model in \gls{piml2}}
The input sequence \mbox{$\mO_t \in \mathbb{R}^{L \times m}$} composed of $L$ observation steps is divided into two parts following~\cite{timeseriesregression}. The first $H$ steps are processed by a transformer encoder $\mathcal{T}_{\text{enc}}$ which captures the vehicle dynamics context and produces a feature representation \mbox{$\mF_{\text{enc}} \in \mathbb{R}^{{H}\times{d}}$}. The decoder $\mathcal{T}_{\text{dec}}$ then  
takes the remaining observations $\mO_{\text{dec}}$ as queries and attends to the encoder output $\mF_{\text{enc}}$ to estimate the vehicle dynamic states $\vz_t \in [\vv_t,\va_t,\boldsymbol{\omega}_t]$, i.\,e., \mbox{$\vz_t = \mathcal{T}_{\text{dec}} (\mO_{\text{dec}},\mF_{\text{enc}}) $} as in Fig.~\ref{fig:informerarchitecture}.



Two extensions were introduced to the transformer-based \gls{ml} model to adapt it for vehicle localization. First, prior knowledge about vehicle dynamic limits is incorporated through a physics guard layer. Second, an uncertainty estimation module was integrated into the model to enable uncertainty-aware predictions that support the subsequent adaptive~\gls{ekf} in vehicle localization.

\subsubsection*{{Physics Guard}}  The domain knowledge about vehicle dynamic limits is embedded in the \gls{ml} model via a physics guard layer to bound the predictions. For example, the maximum attainable acceleration $\va_{\max}$ of a vehicle is governed by the tire-road friction coefficient $\mu$ and bounded by the Kamm's circle:
\begin{equation}
\label{eqn:kamms circle}
a_{\max} = \mu g, \quad \eva_x^2 + \eva_y^2 \leq a_{\max}^2
\end{equation}
Similarly, the vehicle has an upper bound on velocity. The velocity and acceleration estimates from \gls{ml} model are clamped based on their Euclidean norms, while the rotation rate is limited to the maximum allowable angular velocity:  
\begin{equation}
\begin{aligned}
\vu_t &= \tilde{\vu}_t \cdot \min\Big(1, \frac{u_{\max}}{\|\tilde{\vu}_t\|_2 + \epsilon}\Big), 
\\
\boldsymbol{\omega}_t &= \mathrm{clip}(\tilde{\boldsymbol{\omega}}_t, -\boldsymbol{\omega}_{\max}, \boldsymbol{\omega}_{\max}),
\end{aligned}
\end{equation}
where $\tilde{\vu}_t, \tilde{\boldsymbol{\omega}}_t$ denote the raw \gls{ml} predictions, \mbox{$\vu_t \in\{\mu_{\vv},\mu_{\va}\}$} is the clamped velocity or acceleration estimate and $\epsilon$ is a small numerical stabilization term. 
When the physics guard scales a raw prediction, the corresponding variance is scaled by the square of the factor to maintain consistency.
This formulation ensures that the~\gls{ml} predictions remain bounded within the vehicle dynamic limits.

\subsubsection*{{Uncertainty Quantification in~\gls{ml} Model}} The bounded vehicle dynamic estimates $\vz_t$ obtained after the physics guard are approximated as Gaussian random variables~\cite{bishop2006pattern}. 
The \gls{ml} model has a mean $\mu_z[t]$ and a variance $\delta_z[t]$ for each decoder time step after the physics guard. Let \mbox{$\mathcal{D}_t = \{ t \in [\, t\!-\! L\! +\! H\!+\! 1, \, t \,] \}$} denote the set of time steps for $\mathcal{T}_{\text{dec}}$. The predicted mean  and variance  are used to compute the negative log likelihood loss 
\begin{equation}
\label{eqn:nll_loss}
\mathcal{L}_{\text{NLL}}\! = \!\sum_{t \in \mathcal{D}_t}\!
(\vz_t^{\text{gt}}\!-\!\boldsymbol{\mu}_z[t])^{\!\top} 
\!\boldsymbol{\Sigma}_z[t]^{-1} 
(\vz_t^{\text{gt}}\!-\! \boldsymbol{\mu}_z[t])\! 
+\!\log \lvert \boldsymbol{\Sigma}_z[t] \rvert.
\end{equation}

Here, $\boldsymbol{\Sigma}_z[t]$ is the diagonal covariance matrix, with each entry constrained to be strictly positive ($\delta_z[t] > 0$) to ensure numerical stability. This loss formulation allows the \gls{ml} model to express uncertainty by producing higher variance for less certain predictions~\cite{abinav}. During inference, the predicted mean and covariance for vehicle dynamic states at time step $t$ are 
\begin{equation}
\label{equation:inference_ml}
\vz_t = \boldsymbol{\mu}_z[t]\;,\;  \boldsymbol{\Sigma}_z[t] = \operatorname{diag}(\boldsymbol{\delta}_z[t]).
\end{equation}

The proposed approach leverages variance estimates from the \gls{ml} model for the \gls{ekf}, enabling an adaptive measurement noise covariance to improve accuracy. The architecture is lightweight as well, supporting real-time execution.

\subsubsection{\acrfull{ekf} in \gls{piml2}}
\label{sec:ekf}
In the proposed \gls{piml2} architecture, the vehicle dynamic states $\vz_t$ estimated by the \gls{ml} model serve as measurements for the \gls{ekf}. The \gls{ekf}, then recursively updates its belief of the vehicle state $\vx_t$ using these estimates $\vz_t$ and the vehicle motion model. Since the \gls{ml} outputs are fused with a physics-based state-space model implemented via an \gls{ekf}, the overall \gls{piml2} framework is physics-informed.

\subsubsection*{State Model} The \gls{ekf} utilizes a kinematic motion model for robustness against extensive parametrization~\cite{sven}. The state vector of the vehicle $\vx_t$ is defined as: 
\begin{equation}
\label{equation:vehicle state}
\vx_t = [\vd_t,\boldsymbol{\theta}_t,\boldsymbol{\omega}_t,\vv_t,\va_t]^\top,
\end{equation}
where $\vd_t$, $\boldsymbol{\theta}_t$, $\boldsymbol{\omega}_t$, $\vv_t$, and $\va_t$ denote the vehicle’s position, Euler angles, angular velocity, linear velocity, and linear acceleration, respectively. The \textit{a priori} state prediction of the \gls{ekf} is computed using numerical integration under a constant acceleration motion model
\!
\begin{equation}
\label{eq:state_prediction}
\vx_{t+1|t} =
\begin{bmatrix}
\vd_t + \mathbf{R}(\boldsymbol{\theta}_t)\vv_t \Delta t + \tfrac{1}{2}\mathbf{R}(\boldsymbol{\theta}_t)\va_t \Delta t^2 \\
\boldsymbol{\theta}_t + \boldsymbol{\omega}_t \Delta t \\
\boldsymbol{\omega}_t \\
\vv_t + \va_t \Delta t \\
\va_t
\end{bmatrix}
+ \boldsymbol{\eta}_s.
\end{equation}

Here, $\mathbf{R}(\boldsymbol{\theta}_t) \!\in\!SO(3)$ is the rotation matrix computed using roll-pitch-yaw sequence, mapping linear velocities and accelerations from the vehicle frame to the global frame~\cite{botsch}. For typical ground vehicles, small roll and pitch angles ensure that Euler angles do not suffer from gimbal lock. 
The process model in Eq.~(\ref{eq:state_prediction}) is nonlinear due to $\mathbf{R}(\boldsymbol{\theta}_t)$, and its uncertainty is modeled as zero-mean Gaussian noise  \mbox{$\boldsymbol{\eta}_s\!\sim\! \mathcal{N}(\mathbf{0}, \mQ)$}, where $\mQ$ is the process noise covariance.  


\subsubsection*{Measurement Model}
In the proposed EKF, the estimated vehicle dynamic states $\boldsymbol{\mu}_z[t]$ and associated uncertainties $\boldsymbol{\delta}_z[t]$ from the \gls{ml} model serve as pseudo-measurements. The measurement vector and adaptive noise covariance matrix are then defined as
\begin{equation}
\label{eqn:measurement_vector}
\vz_t\! =\! \boldsymbol{\mu}_z[t]\!=\!\big[\mu_{\boldsymbol{\omega}},\mu_{\vv},\mu_{\va}\big]^\top, 
\boldsymbol{\Sigma}_z[t] \!= \!\mathrm{diag}\big(\delta_{\boldsymbol{\omega}},\delta_{\vv},\delta_{\va}\big).
\end{equation}
The \gls{ekf} operates recursively at the sampling frequency of onboard sensor data. The global vehicle pose can be computed using the \textit{a posteriori} estimates $\vx_{t|t}$ of the \gls{ekf} as follows
\begin{equation}
\label{a-posterior}
\mT_t = \mT_{t,0}\mT_0 = \begin{bmatrix} R(\boldsymbol{\theta}[t|t]) & \mathbf{d}[t|t] \\ \mathbf{0}^\top & 1  \end{bmatrix}\mT_0
\end{equation}
where $\mT_{t,0}$ is the cumulative transformation matrix. 


In contrast to many existing vehicle localization pipelines, the EKF component in \gls{piml2} is implemented in a differentiable manner, allowing loss gradients to propagate back to the \gls{ml} model, as detailed in the following subsection.

\subsection{Differentiable Kalman filters as physics regularizer}
\label{sec:KF as physics regularizer}
Conventional \gls{ml} models learn their parameters $\boldsymbol{\Theta}$ purely from data, capturing statistical correlations but not physical relationships or causal dependencies. Consequently, the vehicle dynamic estimates $\vz_t$ in Eq.~(\ref{equation:inference_ml}) may deviate from the physically consistent manifold defined by the vehicle motion models. This lack of physical inductive bias in purely data-driven estimators can be mitigated through physics-based regularization or physics-informed architectures~\cite{physicsinformed}. 
While standard Bayesian filters can impose such constraints at inference, they cannot enforce physics-aware learning of the \gls{ml} model parameters. In \gls{piml2}, the \gls{ekf} is integrated with the \gls{ml} model in a fully differentiable manner, enforcing temporal and physical consistency on the \gls{ml} predictions during training.

\subsubsection{Differentiable \gls{ekf}} The prediction and update steps for the computation of first and second order statistical moments are made fully differentiable to support backpropagation~\cite{kalmannet}. The state prediction step of a standard \gls{ekf}~\cite{botsch} can be defined as
\begin{equation}
\begin{split}
\vx_{t|t-1} &= \mathbf{f}(\vx_{t-1|t-1}) \,, \mF_t = \mathcal{J}_f \big( \mathbf{x}_{t-1|t-1} \big) \\
\mathbf{P}_{t|t-1} &= \mF_t \, \mathbf{P}_{t-1|t-1} \, \mF_t^\top + \mQ.
\end{split}
\end{equation}
Here, $\mathbf{f}$ denotes the nonlinear process model in Eq.~(\ref{eq:state_prediction}), $\mF_t$ is the Jacobian linearization, and $\mQ$ is the process noise matrix. In practice, computing the Jacobian numerically can break the computational graph or introduce instability. To address this, we compute $\mF_t$ analytically as follows:
\!\!
\begin{equation*}
\resizebox{0.45\textwidth}{!}{$
\mF_t\!=\!
\begin{bmatrix}
I_3 & \frac{\partial \mR}{\partial \boldsymbol{\theta}} \left(\!\mathbf{v}_t \Delta t\!+\!\frac{\va_t \Delta t^2}{2} \!\right) & 0_{3\times 3} & \mR \Delta t & \frac{\mR \Delta t^2}{2}  \\
0_{3\times3} & I_3 & \Delta t \, I_3 & 0_{3\times 3} & 0_{3\times 3} \\
0_{3\times 3} & 0_{3\times 3} & I_3 & 0_{3\times 3} & 0_{3\times 3} \\
0_{3\times 3} & 0_{3\times 3} & 0_{3\times 3} & I_3 & \Delta t \, I_3 \\
0_{3\times 3} & 0_{3\times 3} & 0_{3\times 3} & 0_{3\times 3} & I_3
\end{bmatrix}$}
\end{equation*}
Here, $I_3$ represents the $3\!\times\!3$ identity matrix, and $\frac{\partial \mR}{\partial \boldsymbol{\theta}}$ is the standard derivative of the rotation matrix~\cite{botsch}. The observation moments $\hat{\vz}_{t|t-1}, \mS_{t|t-1}$ do not encounter this issue in \gls{piml2}, as the observation matrix $\mH$ corresponds to a direct mapping from the state space to pseudo \gls{ml} measurements. The update step in the \gls{ekf} to compute \textit{a posteriori} state moments is defined as
\begin{equation}
\begin{aligned}
\label{eqn:kalman stable}
\mathbf{x}_{t|t} &= \mathbf{x}_{t|t-1} + \mathbf{K}_t \, (\vz_t -\hat{\vz}_{t|t-1}), \\
\mathbf{P}_{t|t} &= (\mathbf{I}\!-\!\mathbf{K}_t \mathbf{H}) \mathbf{P}_{t|t-1} (\mathbf{I}\!-\!\mathbf{K}_t\mathbf{H})^\top\!\!+\! \mathbf{K}_t \boldsymbol{\Sigma}_z[t] \mathbf{K}_t^\top, \\
\mathbf{K}_t &= \mathbf{P}_{t|t-1} \, \mathbf{H}^\top \, \mathbf{S}_{t|t-1}^{-1}.
\end{aligned}
\end{equation}

Here, the Kalman gain $\mathbf{K}_t$ can suffer from numerical instability due to inversion of the innovation covariance $\mathbf{S}_{t|t-1}$. To address this, $\mathbf{S}_{t|t-1}$ is factorized using Cholesky decomposition, and $\mathbf{K}_t$ is computed by solving a linear system rather than explicit inversion. The posterior covariance $\mathbf{P}_{t|t}$ is updated using the Joseph form, which preserves symmetry and positive semi-definiteness, improving stability for end-to-end training. The differentiable \gls{ekf} additionally supports batch-wise training. 


\subsubsection{Physics-Based Regularization}
This mechanism guides the~\gls{ml} optimization to favour solutions consistent with physical laws, improving generalization~\cite{physicsinformed}. Haarnoja et al.~\cite{backpropkf} first introduced neural network training with a Kalman filter. Subsequent works on differentiable filtering combined classical filtering techniques with learning-based approaches~\cite{particlefilter,iros-diff}. 
However, a systematic analysis of the physical and temporal consistency induced by differentiable filters, particularly in end-to-end training, remains limited.

In contrast, this paper shows that the Kalman filter acts as a physics-based regularizer in end-to-end training. A decomposition of the overall loss function shows that the Kalman filter introduces structured regularization terms that constrain the \gls{ml} parameter space to comply with the underlying physical model. This interpretation is supported by both a formal loss decomposition and empirical validation in \mbox{section~\ref{section:Evaluation and results}} for the vehicle localization task.

Let \mbox{$\mathcal{X}_{\text{gt}} = \{\vx_{0}^{\text{gt}},..,\vx_{t}^{\text{gt}}\}$} denote the sequence of vehicle's ground truth states,  \mbox{$\mathcal{X} = \{\vx_{1|1},..,\vx_{t|t}\}$} denote the \textit{a posteriori} estimates from a differentiable \gls{ekf}, and \mbox{$\mathcal{Z} = \{\vz_{1},..,\vz_{t}\}$} denote the measurements from a \gls{ml} model parameterized by~$\mathbf{\Theta}$. End-to-end training of a \gls{ml} based measurement model \mbox{$g_{\text{ml}}(\vz_t|\mO_t;\boldsymbol{\Theta})$} with a differentiable Kalman filter, by backpropagating the loss through the \textit{a posteriori} estimates, inherently regularizes the learning process. This is motivated by the following mathematical formulation for the loss function:
\begin{equation}
\label{eqn:phy_regularizer}
\begin{aligned}
\mathcal{L}(\mathbf{\Theta}) &= \sum_{t=1}^{T} \big\| \mathbf{x}_{t|t}(\mathbf{\Theta}) - \mathbf{x}_t^{\text{gt}} \big\|^2\\
&= \sum_{t} 
\big\| 
\underbrace{\mathbf{x}_{t|t-1}\!-\! \mathbf{x}_t^{\text{gt}}}_{\text{Prior residual $(\Delta \ve_t)$}} 
\!\!\!+ 
\mathbf{K}_t \underbrace{\big( \mathbf{z}_t - \mH \mathbf{x}_{t|t-1} \big)}_{\text{Innovation ($\vr_t$)}} 
\big\|^2\\
&= \sum_{t} \big\|\mathbf{x}_{t|t-1}(\boldsymbol{\Theta}) - \mathbf{x}_t^{\text{gt}}\big\|^2 + \lambda\,\boldsymbol{\Phi}_t^{\text{KF}}(\boldsymbol{\Theta}).
\end{aligned}
\end{equation}
Here, both the predicted state $\vx_{t|t-1}$ and the Kalman gain $\mathbf{K}_t$ are functions of $\mathbf{\Theta}$, as they are computed recursively from previous $\mathbf{x}_{t-1|t-1}(\mathbf{\Theta})$ and    $\mathbf{z}_t(\mathbf{\Theta})$. Finally, the loss function in Eq.~(\ref{eqn:phy_regularizer}) can be expressed as a standard $L2$ loss $\big\|\Delta\ve_t\big\|^2$, summed over all time steps and a Kalman filter-based regularization term $\boldsymbol{\Phi}_t^{\text{KF}}(\boldsymbol{\Theta})$, defined as 
\begin{equation}
\label{eqn:kf_regularizer}
\boldsymbol{\Phi}_t^{\text{KF}}(\boldsymbol{\Theta})
\!=\!(\mathbf{K}_t \mathbf{r}_t)^\top\!(\mathbf{K}_t \mathbf{r}_t) 
\!+\!2(\Delta \ve_t)^\top\!(\mathbf{K}_t \mathbf{r}_t),\; \lambda = 1
\end{equation}
Thus, learning the parameters $\boldsymbol{\Theta}$ to minimize the Loss in Eq.~(\ref{eqn:phy_regularizer}) can be interpreted as learning the parameters $\boldsymbol{\Theta}$ to minimize the squared prediction error $\Delta \ve_t$ with the regularization term from Eq.~(\ref{eqn:kf_regularizer}). 
The loss encourages the \gls{ml} model to produce measurement estimates consistent with the state model and coherent across time steps.
This understanding emphasizes the interpretation of proposed architecture as a physics-regularized machine learning framework. The loss propagation for \gls{piml2} follows backpropagation through time similar to~\cite{backpropkf}. 
The~\gls{piml2} architecture leverages this \gls{kf}-based physics regularization to improve vehicle localization accuracy and generalization. 

\subsection{Training Strategy for \gls{piml2}}
The training of the \gls{piml2} architecture is divided into two stages to ensure reliable convergence~\cite{pinn_training}.
\subsubsection{Pretraining (Data-Driven)} Initially the \gls{ml} model is pretrained to estimate the vehicle dynamic states $\vz_t$ using a purely data driven approach. The optimization is performed using the negative log-likelihood loss function defined in Eq.~(\ref{eqn:nll_loss}), which jointly improves the mean $\boldsymbol{\mu}_z[t]$ and variance $\boldsymbol{\delta}_z[t]$ estimates. This pretraining stage ensures that the measurement model \mbox{$g_{\text{ml}}(\vz_t|\mO_t;\boldsymbol{\Theta})$} is prepared for subsequent end-to-end training with the differentiable \gls{ekf}.

\subsubsection{Fine-tuning (Physics Regularizer)} The \gls{ml} transformer model is fine-tuned by backpropagating the loss through the differentiable \gls{ekf}. For a single data sample $(\mO_t,\vz^{\text{gt}})$, the \gls{ml} model predicts the measurements~$\vz[t]$ and its covariance matrix~$\boldsymbol{\Sigma}_z[t]$ for all decoder time steps $\mathcal{D}_{t}$. The \gls{ekf} is then executed recursively to produce the \textit{a posteriori} state estimates $\vx_{t|t}$. The overall loss function is defined as  
\begin{equation}
\mathcal{L}_{\text{total}} = \sum_{t \in \mathcal{D}_{t}} \mathcal{L}_{\text{NLL}}
\!+\!\beta\! \cdot\! \tanh\! \left( \!\frac{1}{\beta} \!\sum_{t \in \mathcal{D}_{t}}\! \left\| \mathbf{d}_t\! -\! \mathbf{d}_t^{\text{gt}} \right\|^2 \!\right).
\label{eqn:total_loss_single}
\end{equation}
Here, $\beta$ is a hyperparameter, and $\tanh$ activation is employed to mitigate exploding gradients~\cite{tanh}. The position loss ensures that the estimated vehicle positions $\mathbf{d}_t$ closely align with the ground truth to enhance localization accuracy.

\section{Dataset and Baselines}
\subsection{Dataset}

Most existing dead-reckoning approaches using proprioceptive sensors focus primarily on \gls{imu}-based localization~\mbox{\cite{survey_localization, imu_survey,wheel-imu}}. These methods are typically evaluated on the KITTI~\cite{kitti} dataset or other proprietary datasets~\cite{lstmIMU,sven,dl-avl}. In contrast, this work emphasizes localization using only onboard sensors. The KITTI dataset is not suitable here because it does not provide the necessary onboard sensor data. Therefore, the publicly available \gls{revsted}~\cite{revsted} is used to evaluate the proposed method.

While~\gls{revsted} primarily contains recordings in bright weather and mild rain, evaluating the generalization of physics-regularized models requires more challenging, out-of-distribution conditions. To address this, a novel dataset is introduced under heavy snow and low-friction (low-$\mu$) conditions.
This dataset comprises approximately two hours of driving data (0.37 million samples) collected from multiple drivers to ensure behavioral diversity and includes diverse vehicle dynamic maneuvers. 
Recordings were obtained on a test track using the test vehicle in Fig.~\ref{fig:piml_architecture}. A navigation-grade \gls{imu}~\cite{adma} equipped with RTK positioning provided ground truth vehicle states with centimeter-level accuracy. The datasets provide a comprehensive set of onboard signals $\vo_t$ at 50 Hz, as described below.
\begin{equation}
\label{equation:obd_data}
\vo_t = \begin{bmatrix}
v_\text{s}, \delta_{\text{sw}}, \dot{\psi}, a_{\text{y}}, p_{\text{br}}, v_{\text{fl}}, v_{\text{fr}}, v_{\text{rl}}, v_{\text{rr}}
\end{bmatrix}^\mathrm{T},
\end{equation}
where $v_{\text{s}}$ denotes speedometer reading, $\delta_{\text{sw}}$ the steering wheel position, $\dot{\psi}$ the yaw rate, \mbox{$v_{\text{fl}}, v_{\text{fr}}, v_{\text{rl}}, v_{\text{rr}}$} the wheel speeds, $p_{\text{br}}$ the brake pressure, and $a_{\text{y}}$ the lateral acceleration. 

As the onboard sensors do not provide height measurements, the vehicle motion is assumed to be planar. Accordingly, localization is performed in 2D using the estimated vehicle dynamic states defined at the vehicle's center of gravity (COG) in the vehicle frame,
\mbox{$\vz_t = [\evv_x, \evv_y, \eva_x, \eva_y, \omega_z]^\mathrm{T}.$}

\subsection{Baselines}


The proposed method is compared against the following baseline methods. 
These baselines estimate the vehicle dynamic states $\vz_t$ using \acrfull{osd}. 
Vehicle localization is then obtained by processing $\vz_t$ as pseudo measurements in a standard \gls{ekf} (for baselines 1-4), analogous to the differentiable \gls{ekf} in \mbox{section~\ref{sec:KF as physics regularizer}}.
The process and measurement noise matrices for these EKF models are tuned using the validation dataset. Additionally, an advanced baseline~(5), using joint training with \gls{ekf} is also compared. 

\subsubsection{\acrshort{osd}-Baseline} 
This baseline serves as a simple classical model-based approach. 
Unlike \gls{imu}-based localization, classical motion models cannot be readily implemented using \gls{osd} due to the lack of longitudinal acceleration $a_{\text{x}}$ data in \gls{revsted}~\cite{abinav}.
However, wheel speed measurements allow an approximate estimation of the longitudinal velocity $v_x$ for a simple kinematic model~\cite[p.~414]{isermann}. 
The resulting baseline estimates are given by:
\begin{equation}
\begin{aligned}
\label{eqn:obd}
\evv_x\!&\approx\!\frac{v_{\text{fl}}\!+\!v_{\text{fr}}\!+\!v_{\text{rl}}\!+ \!v_{\text{rr}}}{4}, \eva_y\!=\!a_{y}^{\text{osd}}\!-\!b_{a_y},
\omega_z\!=\!\omega_z^{\text{osd}}\!-\!b_{\omega_z}
\end{aligned}
\end{equation}
where $b_{a_y}$ and $b_{\omega_z}$ denote the standstill sensor biases~\cite{sven}.

\subsubsection{NN-VDM~\cite{nn-vdm}} A neural network-based vehicle dynamics model employing a \gls{gru} architecture for dynamic state estimation. 
\subsubsection{DL-AVL~\cite{dl-avl}} A \gls{lstm}-based model that refines noisy~\gls{imu} data for localization.
\subsubsection{RNN-EKF~\cite{rnn-ekf}} A complex architecture combining a \gls{gru} network with attention to improve localization.
\subsubsection{Backprop-KF~\cite{backpropkf}} An observation model trained jointly with a differentiable \gls{ekf}. The original learnt model is adapted to suit~\gls{osd}. 



Although the architectures in~\cite{rnn-ekf,dl-avl} were developed for IMU-based localization, their problem formulation closely aligns with this work, making them suitable baselines.


\section{Experiments and results}
This section presents the experimental setup, implementation details, and evaluation results. It aims to address the following research questions: (1)~How accurately does \gls{piml2} estimate the vehicle dynamic states $\vz_t$? (2)~What performance gains does a physics-regularized machine learning architecture provide for vehicle localization? (3)~Does \gls{piml2} have improved generalization to low-$\mu$ conditions? (4)~Can the proposed architecture operate in real-time? 


\subsection{Experimental Setup}

\subsubsection{Implementation details} 

The transformer-based \gls{ml} model uses two attention layers with eight heads (d = 512), with \mbox{$L=250$} and \mbox{$H=150$}.
The \gls{ml} model and differentiable \gls{ekf} are trained jointly using Adam (\mbox{$lr=10^{-4}$}, cosine schedule,~\mbox{$b=250$}) for 20 epochs.
Pretraining and fine-tuning in \gls{piml2} were split in a \mbox{$70\!:\!30$} ratio of total epochs to ensure reliable convergence. 
The total loss scaling parameter was set to \mbox{$\beta=8$} based on a hyperparameter study. 
The physics guard imposes a friction-dependent acceleration bound described in Eq.~(\ref{eqn:kamms circle}), yielding \mbox{$a_{\text{max}}=\SI{10}{\meter\per\second\squared}$} for dry road and \mbox{$a_{\text{max}}=\SI{5}{\meter\per\second\squared}$} for low-friction conditions ($\mu \leq 0.5$).  The velocity and yaw-rate limits are fixed at \mbox{$v_{\text{max}}=\SI{30}{\meter\per\second}$} and \mbox{$\omega_{\text{max}} = \SI{1.5}{\radian\per\second}$}, respectively.
The \gls{revsted} dataset is partitioned into training, validation, and test in a 70:10:20 ratio, ensuring temporal continuity across the time domain as proposed in~\cite{abinav}. 
All training was performed on a single NVIDIA Quadro RTX 5000 GPU.

\subsubsection{Evaluation metrics}
The accuracy of vehicle dynamic state estimation is evaluated using the widely adopted \gls{rmse}~\cite{nn-vdm} metric. Vehicle localization accuracy during dead reckoning is assessed using \gls{rmse} and \gls{me} metrics~\cite{sven,wheel-imu}. 



\subsection{Evaluation and Results}
\label{section:Evaluation and results}
\subsubsection{Evaluation of vehicle dynamic state estimation} 
The proposed \gls{piml2} architecture outperformed all five baselines, as well as \gls{piml2}*, which corresponds to training the \gls{ml} component for total epochs without the fine-tuning stage. A quantitative evaluation on the \gls{revsted} test dataset is presented in Table~\ref{tab:1}.  Compared to the previous best performing baseline model \mbox{NN-VDM~\cite{nn-vdm}} for $\vz_t$ estimation, the proposed~\gls{piml2} architecture achieves significant improvements in $\evv_x$ and $\omega_z$ estimation. This gain can be attributed to both the attention mechanism in the transformer model and physics-based regularization introduced by the differentiable \gls{ekf}. Notably, the 28\% and 25\% improvement of \gls{piml2} over the pretraining variant in  $\evv_x$ and $\omega_z$ estimation highlights the effectiveness of \gls{ekf} based physics regularization during the fine-tuning stage. 


This shows that \gls{piml2} improves vehicle dynamic state estimates, providing more reliable inputs for localization. 
\vspace{-0.3em}
\begin{table}[htbp]
\centering
\caption{Comparison of vehicle dynamic state estimation using \gls{rmse} in \gls{revsted}~\cite{revsted}}
\begin{tabular}{lccccc}
\toprule
\textbf{Models} & $\evv_x$ & $\evv_y$ & $\eva_x$ & $\eva_y$ & $\omega_z$ \\ 
\midrule
OSD-Baseline & 0.22 & - & - & 0.20 & 0.43\\ 
NN-VDM~\cite{nn-vdm} & 0.06 & 0.03 & \textbf{0.13} & 0.13 & 0.41 \\ 
DL-AVL~\cite{dl-avl} & 0.07 & 0.03 & 0.16 & 0.13 & 0.46 \\ 
RNN-EKF~\cite{rnn-ekf} & 0.07 & 0.02 & \textbf{0.13} & 0.13 & 0.44 \\ 
Backprop-KF~\cite{backpropkf} & 0.12 & 0.02 & 0.31 & 0.15 & 0.38 \\ 
\midrule
\gls{piml2}* & 0.03 & \textbf{0.01} & 0.18 & 0.13 & 0.24 \\ 
\gls{piml2} & \textbf{0.02} & \textbf{0.01} & 0.15 & \textbf{0.12} & \textbf{0.18} \\ 
\bottomrule
\end{tabular}%
\begin{flushleft}
\footnotesize ~\gls{piml2}* denotes the variant without physics-based regularization.\\
\footnotesize ~\gls{osd}-Baseline cannot estimate dashed quantities.
\end{flushleft}
\label{tab:1}
\end{table}

\begin{figure*}[t]  
\begin{flushleft} 

    \begin{subfigure}[b]{0.24\textwidth}  
        \centering
        \includegraphics[width=\linewidth]{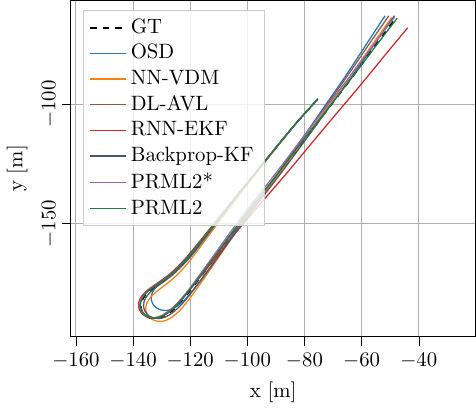}
        \caption{Longitudinal}
        \label{fig:drift1}
    \end{subfigure}
    \hspace{-8pt}
    \begin{subfigure}[b]{0.24\textwidth}
        \centering
        \includegraphics[width=\linewidth]{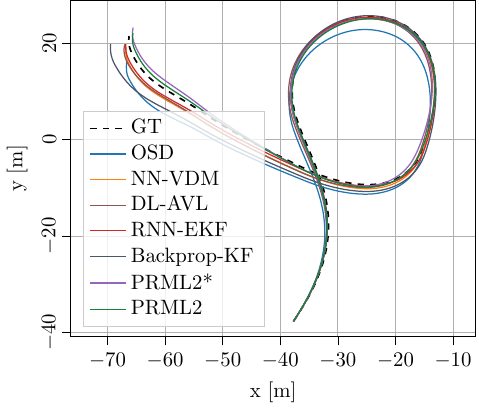}
        \caption{Lateral}
        \label{fig:drift2}
    \end{subfigure}
    \hspace{-8pt}
    \begin{subfigure}[b]{0.24\textwidth}
        \centering
        \includegraphics[width=\linewidth]{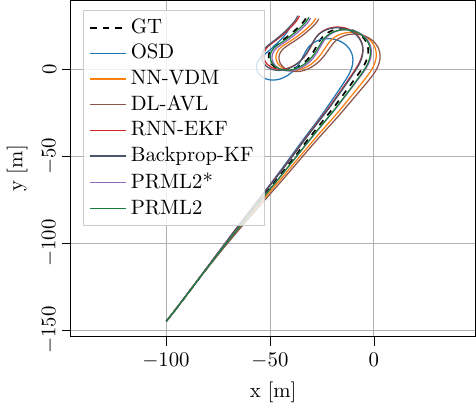}
        \caption{Mixed}
        \label{fig:drift3}
    \end{subfigure}
    \hspace{-8pt}
    \begin{subfigure}[b]{0.24\textwidth}
        \centering
        \includegraphics[width=\linewidth]{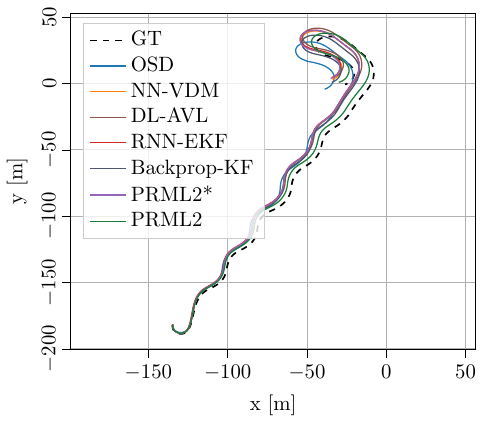}
        \caption{{Snow (low-$\mu$)}}
        \label{fig:drift4}
    \end{subfigure}
    \caption{Vehicle localization performance over \SI{60}{\second} under longitudinal, lateral, and mixed dynamic conditions in the \gls{revsted} dataset. The Snow (low-$\mu$) indicates generalization performance on the proposed low-friction dataset without training.}
    \label{fig:drift_over_time_all_models}
    \vspace{-2em}
    \end{flushleft}
\end{figure*}
\vspace{-0.5em}
\subsubsection{Accuracy of vehicle localization using onboard sensors} 
The onboard sensor-based localization was evaluated over \mbox{$\SI{60}{\second}$} intervals, a standard approach for assessing drift during GNSS outages~\cite{localisation_survey,rnn-ekf}. To ensure a robust evaluation, 45 non-overlapping one-minute scenarios were analyzed in \gls{revsted}, and the aggregated results are summarised in Table~\ref{tab:2}. The proposed \gls{piml2} architecture achieved the highest localization accuracy among all compared methods. Compared to \mbox{RNN-EKF~\cite{rnn-ekf}}, the position \gls{rmse} was reduced by~28\%. The \gls{me} metric representing the worst-case position drift within each \mbox{$\SI{60}{\second}$} sequence was also lowest for \gls{piml2}, highlighting its reliability.


\begin{table}[htbp]
\centering
\caption{Horizontal position error (\si{\meter}) and its standard deviation over \SI{60}{\second} dead-reckoning using onboard sensors.}
\begin{tabular}{l@{}cc|cc}
\toprule
 & \multicolumn{2}{c}{\textbf{\gls{revsted}}} & \multicolumn{2}{c}{\textbf{Snow (low-$\mu$)}} \\ 
\cmidrule(lr){2-3} \cmidrule(lr){4-5}
\textbf{Models} & \textbf{\gls{rmse}} & \textbf{\gls{me}} & \textbf{\gls{rmse}} & \textbf{\gls{me}} \\ 
\midrule
\gls{osd}-Baseline & $5.3\pm2.1$ & $9.2\pm4.0$ & $5.7\pm3.1$ & $9.7\pm4.8$ \\ 
NN-VDM~\cite{nn-vdm} & $2.9\pm2.4$ & $5.6\pm4.7$ & $3.7\pm2.7$ & $7.2\pm5.0$ \\ 
DL-AVL~\cite{dl-avl} & $2.5\pm2.2$ & $5.0\pm4.1$ & $3.8\pm2.7$ & $7.5\pm5.0$ \\ 
RNN-EKF~\cite{rnn-ekf} & $2.1\pm1.5$ & $4.1\pm3.1$ & $3.8\pm2.6$ & $7.3\pm4.9$ \\ 
Backprop-KF~\cite{backpropkf} & $2.6\pm1.5$ & $5.2\pm3.4$ & $3.4\pm2.3$ & $6.0\pm4.1$\\ 
\midrule
\gls{piml2}* & $2.2\pm1.9$ & $4.2\pm3.6$ & $3.6\pm2.7$ & $7.1\pm4.9$ \\ 
\gls{piml2} &  $\mathbf{1.5}\pm\mathbf{1.2}$ & $\mathbf{2.8}\pm\mathbf{2.2}$ & $\mathbf{2.5}\pm\mathbf{1.8}$ & $\mathbf{4.3}\pm\mathbf{3.1}$\\
\bottomrule
\end{tabular}
\begin{flushleft}
\footnotesize  The Snow (low-$\mu$) dataset evaluates generalization performance.
\end{flushleft}
\label{tab:2}
\vspace{-2em}
\end{table}


While the \mbox{RNN-EKF~\cite{rnn-ekf}} matches the pretraining-only variant \gls{piml2}* in \gls{revsted}, adding physics-based regularization via \gls{ekf} improves performance. The Backprop-KF~\cite{backpropkf} underperforms, likely due to its limited observation model. Unlike prior differentiable filters~\cite{backpropkf,iros-diff}, which rely mainly on recursive filtering for temporal structure, \gls{piml2} models it using both the \gls{ml} model and the filter.

\begin{figure}[htbp]
    \centering

    \begin{subfigure}[b]{0.49\linewidth}
        \centering
        \includegraphics[width=\linewidth]{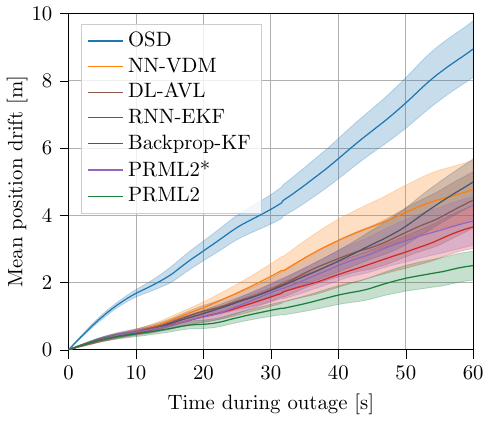}
        \caption{Mean position drift.}
        \label{fig:drift_over_time}
    \end{subfigure}
    \hfill
    \begin{subfigure}[b]{0.49\linewidth}
        \centering
        \includegraphics[width=\linewidth]{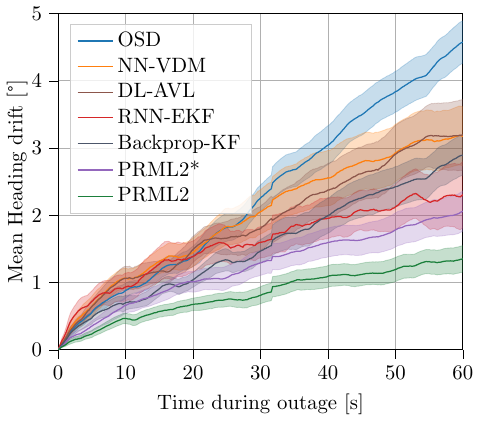}
        \caption{Mean heading drift.}
        \label{fig:another_metric}
    \end{subfigure}

    \caption{Drift evolution during dead-reckoning in \gls{revsted}. The shaded region represents the $\pm0.25\sigma$ band.}
    \label{fig:mean drift over time}
\end{figure}

Qualitative localization results across different vehicle dynamic maneuvers in ReV-StED (Fig.~\ref{fig:drift_over_time_all_models}) demonstrate the localization accuracy and temporal consistency of the proposed approach.
An analysis of the drift rate during dead-reckoning further confirms this advantage, with PRML2 exhibiting the lowest error accumulation over time, as shown in Fig.~\ref{fig:mean drift over time}. The proposed \gls{piml2} achieves a translational error of 1.07\% over 60s intervals on ReV-StED, showing reliable short-term proprioceptive localization using only the onboard sensors.

\subsubsection{Generalization of localization performance} 
The models trained on the \gls{revsted}~\cite{revsted} are evaluated on the proposed low-$\mu$ dataset to test generalization. While some degradation is expected in open-set scenarios, \gls{piml2} achieves the highest localization accuracy under low-$\mu$ conditions, with the lowest mean \gls{rmse} and standard deviation across 45 scenarios (Table~\ref{tab:2}).
The Backprop-KF~\cite{backpropkf}, incorporating an \gls{ekf} in its training regime shows the second-best localization performance.
These results empirically support that the \gls{kf} acts as a physics-based regularizer during end-to-end training, guiding the learning process to improve physical and temporal consistency while enhancing generalization.

Robust localization under unseen and challenging conditions is crucial for safety-critical applications such as autonomous driving. The ability of \gls{piml2} to generalize well to low-$\mu$ scenarios highlights its robustness and practical applicability. 

\subsubsection{Assessment of real-time capability} 
The \gls{piml2} architecture was evaluated for real-time deployment on an NVIDIA Orin edge device~\cite{nvidia_jetson_orin}. The proposed approach has an inference latency of \mbox{$25$--$32~\si{\milli\second}$}, enabling operation at approximately \mbox{$30~\si{\hertz}$} for real-time vehicle localization. 
This highlights that \gls{piml2}, relying solely on onboard sensors, is capable of high-frequency real-time operation.

\subsection{Ablation Study}
An ablation study was conducted to assess the individual contributions of the \gls{piml2} components, with results summarized in Table~\ref{tab:ablation-study}. Incorporating uncertainty estimation into the \gls{ml} model improves localization performance. The physics guard has a minimal effect as it was activated in less than 2\% of predictions. Nevertheless, its integration guarantees strict compliance with vehicle dynamic bounds, preventing physically implausible outputs.

\begin{table}[htbp]
\caption{\gls{piml2} with selective components in \gls{revsted}.}
\centering
\begin{tabular}{ccccc}
\toprule
Physics guard & Uncertainty & EKF (Reg) & \gls{rmse}  & \gls{me}\\
\midrule
 &  & \checkmark &  1.7 & 2.9 \\
\midrule
 & \checkmark & \checkmark & 1.5 & 2.9\\
\midrule
\checkmark & \checkmark & \checkmark &  1.5 & 2.8\\
\bottomrule
\end{tabular}
\label{tab:ablation-study}
\end{table}

\section{Conclusion}
The proposed \gls{piml2} architecture enables accurate estimation of vehicle dynamic states using only onboard proprioceptive sensors. Its physics-informed design incorporates \textit{a priori} domain knowledge and uncertainty modelling to improve vehicle pose estimation. 
We show that Differentiable Kalman filters inherently act as physics-based regularizers, promoting physical and temporal consistency in \gls{ml}-based measurement models. 
\gls{piml2} exploits this property to consistently outperform the compared onboard sensor-based localization methods. 
The physics-based regularization further improves generalization to low-friction driving conditions, which is crucial for autonomous mobility systems. These findings demonstrate that \gls{piml2} offers a robust, real-time solution for proprioceptive vehicle localization using only readily available onboard sensors.



In this work, \gls{piml2} is trained offline and evaluated for planar (2D) localization. Future work will explore extensions to 3D with additional sensors, transfer learning across vehicles and robotic platforms, and online adaptation during periods of satellite signal availability.


\section{ACKNOWLEDGMENT}
This work was funded by the Deutsche Forschungsgemeinschaft (DFG, German Research Foundation) – FIP 135/1, no. 549102058, and by the Federal Ministry of Education and Research of Germany (BMBF) – HyMne2, no. 13FH7I13IA. The authors would like to thank GeneSys Elektronik GmbH and Gaurav Yadav for their support.

\addtolength{\textheight}{-12cm}   





\bibliographystyle{IEEEtran}
\bibliography{root}

\end{document}

%% file: gls.tex
\newacronym{adas}{ADAS}{Advanced Driver Assistance Systems}
\newacronym{gnss}{GNSS}{Global Navigation Satellite System}
\newacronym{piml2}{PRML2}{Physics-Regularized Machine Learning for Localization}
\newacronym{kf}{KF}{Kalman Filter}
\newacronym{ekf}{EKF}{Extended Kalman Filter}
\newacronym{rmse}{RMSE}{Root Mean Squared Error}
\newacronym{gru}{GRU}{Gated Recurrent Unit}
\newacronym{aeb}{AEB}{Autonomous Emergency Braking}
\newacronym{esc}{ESC}{Electronic Stability Control}
\newacronym{mems}{MEMS}{Micro-Electro-Mechanical System}
\newacronym{vsa}{VSA}{Vehicle Sideslip Angle}
\newacronym{ml}{ML}{Machine Learning}
\newacronym{ai}{AI}{Artificial Intelligence}
\newacronym{vmm1}{VMM-1}{Vehicle Motion Model - 1}
\newacronym{vmm2}{VMM-2}{Vehicle Motion Model - 2}
\newacronym{vmm}{VMM}{Vehicle Motion Model}
\newacronym{vmms}{VMMs}{Vehicle Motion Models}
\newacronym{uahi}{UAHL}{Uncertainty-Aware Hybrid Learning}
\newacronym{obd}{OBD}{On Board Diagnostics}
\newacronym{ros}{ROS}{Robot Operating System}
\newacronym{can}{CAN}{Controller Area Network}
\newacronym{pid}{PID}{Parameter IDentification}
\newacronym{gps}{GPS}{Global Positioning Systems}
\newacronym{lstm}{LSTM}{Long Short-Term Memory}
\newacronym{cg}{COG}{Centre of Gravity}
\newacronym{gmm}{GMM}{Gaussian Mixture Model}
\newacronym{revsted}{ReV-StED}{Real-world Vehicle State Estimation Dataset}
\newacronym{sota}{SOTA}{State-of-the-art}
\newacronym{imu}{IMU}{Inertial Measurement Unit}
\newacronym{ins}{INS}{Inertial Navigation System}
\newacronym{dgnss}{DGNSS}{Differential Global Navigation Satellite System}
\newacronym{irs}{IRS}{Inertial Reference System}
\newacronym{adma}{ADMA}{Automotive Dynamic Motion Analyzer}
\newacronym{me}{MAX}{Max Error}
\newacronym{mse}{MSE}{Mean Squared Error}
\newacronym{mae}{MAE}{Mean Absolute Error}
\newacronym{ptp}{PTP}{Precision Time Protocol}
\newacronym{poi}{PoI}{Point of Installation}
\newacronym{cog}{CoG}{Center of Gravity}
\newacronym{gmr}{GMR}{Gaussian Mixture Regression}
\newacronym{osd}{OSD}{Onboard Sensor Data}

%% file: math_commands.tex
\def\eqref#1{equation~\ref{#1}}

\def\1{\bm{1}}

\def\va{{\bm{a}}}

\def\vd{{\bm{d}}}
\def\ve{{\bm{e}}}

\def\vo{{\bm{o}}}

\def\vr{{\bm{r}}}

\def\vu{{\bm{u}}}
\def\vv{{\bm{v}}}

\def\vx{{\bm{x}}}

\def\vz{{\bm{z}}}

\def\eva{{a}}

\def\evv{{v}}

\def\mF{{\bm{F}}}

\def\mH{{\bm{H}}}

\def\mO{{\bm{O}}}

\def\mQ{{\bm{Q}}}
\def\mR{{\bm{R}}}
\def\mS{{\bm{S}}}
\def\mT{{\bm{T}}}

\DeclareMathAlphabet{\mathsfit}{\encodingdefault}{\sfdefault}{m}{sl}
\SetMathAlphabet{\mathsfit}{bold}{\encodingdefault}{\sfdefault}{bx}{n}

